\documentclass[11pt]{article}

\usepackage{acl}

\usepackage{enumitem}
\usepackage{times}
\usepackage{latexsym}

\usepackage{bbm}

\usepackage{amsmath}
\usepackage{amssymb}

\usepackage{booktabs}

\usepackage[most]{tcolorbox}
\usepackage{xcolor}
\usepackage{colortbl}
\usepackage{enumitem}

\definecolor{graybg}{gray}{0.95} 

\usepackage[T1]{fontenc}

\usepackage[utf8]{inputenc}

\usepackage{microtype}

\usepackage{inconsolata}

\usepackage{graphicx}

\title{StratMem-Bench: Evaluating Strategic Memory Use in Virtual\\ Character Conversation Beyond Factual Recall}

\author{
  Yerong Wu\textsuperscript{1,2},
  Tianxing Wu\textsuperscript{1,2}\thanks{Corresponding authors.},
  Minghao Zhu\textsuperscript{3},
  Hangyu Sha\textsuperscript{1,2},
  \textbf{Haofen Wang\textsuperscript{4}\footnotemark[1]} \\
\textsuperscript{1}School of Computer Science and Engineering, Southeast University, China\\ 
\textsuperscript{2}Key Laboratory of New Generation Artificial Intelligence Technology and\\ its Interdisciplinary Applications (Southeast University), Ministry of Education, China\\
\textsuperscript{3}Independent Researcher\\
\textsuperscript{4}College of Design and Innovation, Tongji University, China\\
\texttt{\{yerong.wu,tianxingwu\}@seu.edu.cn, carter.whfcarter@gmail.com} \\
}

\begin{document}
\maketitle
\begin{abstract}
Achieving realistic human-like conversation for virtual characters requires not only a simple memorization and recall of past events, but also the strategic utilization of memory to meet factual needs and social engagement. Current memory utilization relevant (e.g., memory-augmented generation, long-term dialogue, and etc.) benchmarks overlook this nuance, treating memory primarily as a static repository of facts rather than a dynamic resource to be strategically deployed in dialogues. To address this gap, we design \textsc{StratMem-Bench}, a new benchmark to evaluate strategic memory use in character-centric dialogues. This dataset comprises 657 instances where virtual characters must navigate heterogeneous memory pools containing required, supportive, and irrelevant memories. We also propose a framework with different evaluation metrics including Strict Memory Compliance, Memory Integration Quality, Proactive Enrichment Score and Conditional Irrelevance Rate, and  to evaluate strategic memory use capabilities of virtual characters. Experiments on \textsc{StratMem-Bench} which leverage 
the state-of-the-art large language models as virtual characters show that all models perform well at distinguishing between required and irrelevant memories, but struggle once supportive memories are introduced into the decision process.

\end{abstract}

\section{Introduction}
\label{sec:introduction}
\begin{figure*}[t]
  \centering
  \includegraphics[width=\textwidth]{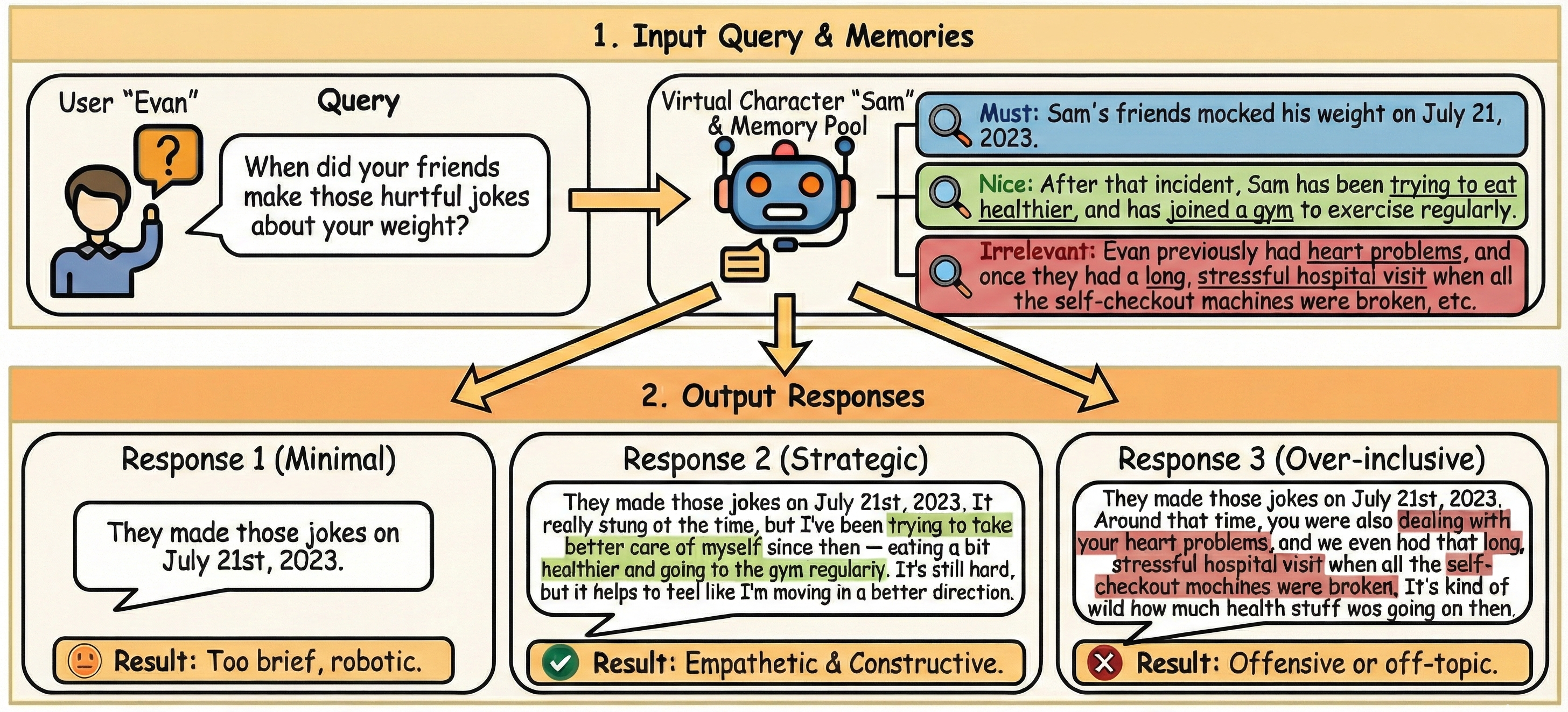}
  \caption{An example of strategic memory use.
  \textbf{Top:} The input contains a user query and memories categorized by different functional roles: \textbf{must} (required), \textbf{nice} (supportive), and \textbf{irr} (irrelevant).
  \textbf{Bottom:} The comparison of the responses obtained by different memory use strategies. Response~1 exhibits the results of \emph{minimal} memory use, relying only on \textbf{must} memories, Response~2 exhibits the results of \emph{strategic} memory use combining \textbf{must} and \textbf{nice} memories, and Response~3 exhibits the results of \emph{over-inclusive} memory use incorporating \textbf{irr} memories.}
  \label{fig:sam-example}
\end{figure*}

Memory use in human conversation involves more than the retrieval of stored information, and speakers must strategically decide which information to deploy in response to the demands of an interaction. Information aligned with a speaker’s current dialogue objective is selectively activated via spreading activation mechanisms \citep{collins1975spreading, anderson1983architecture}, whereas information that does not serve the ongoing communicative context is actively suppressed \citep{anderson2001suppressing, nelson1990metamemory}. Thus, conversational utterances emerge from a continual process of selection, in which speakers determine what to express based on its relevance to their current social goals \citep{pasupathi2001social, alea2003thinking}. This notion is consistent with pragmatic theories of communication, such as the \emph{Gricean Maxims} \citep{grice1975logic}, which require speakers to provide information that is both relevant to the conversational context (Maxim of Relation) and appropriately informative (Maxim of Quantity). 
To enable virtual characters to speak in a human-like manner, they must exhibit strategic memory control analogous to that observed in human conversation. In recent years, large language models (LLMs) have become the dominant backbone for building virtual characters \citep{maharana2024locomo}, owing to their strong capabilities of text generation and context understanding. However, despite having access to the given memories, modern LLM-based virtual characters lack the ability of strategically using this information for generating responses \citep{wu2024longmemeval, maharana2024locomo}.




To solve this problem, it is necessary to make strategic memory use measurable. We first partition memories into three categories, i.e., \emph{required}, \emph{supportive}, and \emph{irrelevant} memories denoted as \textbf{must}, \textbf{nice}, and \textbf{irr}, respectively. \textbf{must} memories are \emph{required} to ensure response correctness; \textbf{nice} memories provide \emph{supportive} information which can enrich responses through personalization, empathy, or social coherence; \textbf{irr} memories correspond to information that is \emph{irrelevant} to the current interaction and should be actively suppressed. Based on this, strategic memory use refers to using all \textbf{must} memories, incorporating some \textbf{nice} memories in appropriate ways, and avoiding the use of \textbf{irr} memories. Figure~\ref{fig:sam-example} illustrates an example of strategic memory use, comparing it with minimal memory use (with only \textbf{must} memories) and over-inclusive memory use (introducing \textbf{irr} memories).

Based on this classification, we then construct a new benchmark called \textsc{StratMem-Bench} to evaluate strategic memory use in virtual character conversation. \textsc{StratMem-Bench} consists of 657 instances, each of which corresponds to the dialogue history, the current user query, a persona, and memories covering the three types introduced above. The test model is required to select and integrate memories for generating responses. We finally propose four new evaluation metrics to measure whether all \textbf{must} memories are used, whether some \textbf{nice} memories are appropriately incorporated, whether \textbf{irr} memories are avoided, and the quality of integrating selected memories, for all instances.

Under this evaluation setting, our design fundamentally differs from previous works (e.g., \citealp{yang2024crag, maharana2024locomo, wu2024longmemeval, rau2024bergen}). Existing memory-oriented evaluation paradigms for memory-augmented generation, long-term dialogue, and related tasks primarily focus on evaluating factual recall, which typically emphasizes whether retrieved factual information is relevant to a given query and reflected in the generated response. A good factual recall therefore corresponds to using as many \textbf{must} memories as possible in our setting. However, this is insufficient in character-centric dialogues, where response quality depends not only on correctly using \textbf{must} memories, but also on selectively incorporating \textbf{nice} memories while avoiding \textbf{irr} memories.

In summary, our contributions are three-fold:

\begin{itemize}[leftmargin=*]
\item We design \textsc{StratMem-Bench}, the first benchmark to explicitly distinguish between required, supportive, and irrelevant memories in character-centric dialogues, thereby expanding the evaluation beyond factual recall to strategic memory use in response generation.
\item We propose a comprehensive evaluation framework for strategic memory use, introducing four new evaluation metrics, Strict Memory Compliance, Memory Integration Quality, Proactive Enrichment Score, and Conditional Irrelevance Rate to capture different aspects of models' capabilities on selecting and integrating memories for response generation.  
\item We conduct a systematic analysis of the state-of-the-art LLMs taken as virtual characters for strategic memory use, and there is no evaluated LLM demonstrating an obvious advantage. LLMs are skilled at handling \textbf{must} memories, but struggle to maintain a balanced use of \textbf{nice} and \textbf{irr} memories.
\end{itemize}


\section{Related Work}
\subsection{LLM-based Virtual Characters} 
Existing studies have demonstrated that LLMs can perform role-playing effectively when conditioned on persona descriptions, character backstories, or instruction-tuned prompts, enabling the simulation of virtual characters with coherent and recognizable behaviors \cite{wang2024rolellm, shao2023characterllm, ng2024echo}. Relevant benchmarks and evaluation frameworks primarily evaluate role-play quality in terms of static character consistency \cite{tu2024charactereval, samuel2024personagym}. Recent works have begun to move beyond static persona consistency by studying LLM-based role-playing in longer interactive settings with evolving conversational contexts \cite{maharana2024locomo, he2025madial}. 
Under this trend, the ability to retain and utilize long-term memories becomes increasingly indispensable.
However, these works do not consider the need of strategic memory use in virtual character conversation for making turn-level decisions (i.e., selecting and integrating memories in response to the current query), so there is no corresponding evaluation benchmark.

\subsection{Memory Utilization Relevant Benchmarks}
Prior works \citep{rau2024bergen, hengle2025multilingual, wu2024longmemeval, yang2024crag, he2025madial} on memory-augmented generation, long-term dialogue, and other tasks have introduced a variety of benchmarks for evaluating the performance of memory utilization. For example, CRAG \citep{yang2024crag} provides a comprehensive benchmark for retrieval-augmented generation (RAG) \citep{gao2023retrieval,zhao2024awecita}, evaluating memory-enhanced question answering across multiple domains and diverse question types. 
LoCoMo \citep{maharana2024locomo} evaluates whether models can retrieve effective memories across dialogue sessions for cross-session question answering and event summarization over very long conversations. LongMemEval \citep{wu2024longmemeval} evaluates chat assistants' ability of not only retaining and retrieving memories from prior user–assistant interactions across dialogue sessions, but also reasoning over such retrieved memories.

All of the above benchmarks focus on evaluating factual recall, where memory utilization is primarily assessed by whether memory including specific objective facts (e.g., time, location, and entity names) can be accurately retrieved and reflected in the response to fulfill user's request.
While effective for measuring factual recall, such an evaluation setting is insufficient for human-like virtual character conversation, where response quality depends on accessing required memories, as well as how models strategically select and integrate memories during generation.



\subsection{Personalized RAG}
Personalized RAG aims to enrich response generation by incorporating user-specific information, such as preferences, history, or contextual signals, to better adapt outputs to individual users \citep{li2025survey}. Existing benchmarks in this area mainly evaluate whether generated responses appropriately reflect personalized user information \citep{wang2024crafting, shi2025cfrag, jiang2025know, zhao2025personalens, zhao2024prefeval}. 
Although personalized RAG also recognizes the selective use of available user information, it does not rely on the LLM for generation itself but some other pre-designed methods. In other words, prior works in personalized RAG do not explicitly model the strategic use of information during generation, which is the main focus of our study. Thus, existing benchmarks for personalized RAG cannot be used to evaluate strategic memory use in character-centric dialogue scenarios.

\section{StratMem-Bench}
\label{sec:StratMem-Bench}

\subsection{Task Definition}
\label{sec:task-formulation}
We evaluate \emph{strategic memory use} through a task of conditional response generation:
\begin{equation}
\hat{y} = f(h, q, M, P)
\end{equation}
where each input instance $x_i = (h_i, q_i, P_i, M_i)$ is composed of a dialogue history $h_i$ containing all utterances up to the current turn from both the user and the virtual character, the current user query $q_i$, a memory set $M_i = \{m_1, m_2, \ldots, m_{|M_i|}\}$ extracted from previous interactions, and a persona $P_i$ that specifies the virtual character's traits, background, preferences, and values.

To distinguish memories in character-centric dialogues, we annotate the memory pool $M_i$ with functional roles at the instance level. Specifically, memory items are partitioned into three disjoint subsets: $M_{i,\text{must}}$, $M_{i,\text{nice}}$, and $M_{i,\text{irr}}$, representing memories that are required, supportive, and irrelevant for generating an appropriate response, respectively.

Importantly, these annotations are \emph{not} accessible to the model during response generation. Instead, the model receives $M_i$ as an unlabeled set of memory items presented in a uniform natural-language format, without any explicit role indicators. Since memory roles are not provided and no explicit instructions are given regarding which types of memories to prioritize, models cannot rely on instruction following to solve the task. Moreover, these annotations are instance-specific, meaning that the role of a memory item depends on the current query and dialogue context, rather than being a fixed property of the memory itself. As a result, models must implicitly infer the functional relevance of each memory item in context and select appropriate memories for producing the response $\hat{y}_i$, i.e., strategic memory use beyond factual recall.

\subsection{Dataset Construction}
\label{sec:dataset}
\begin{figure}[t]
  \centering
  \includegraphics[width=\columnwidth]{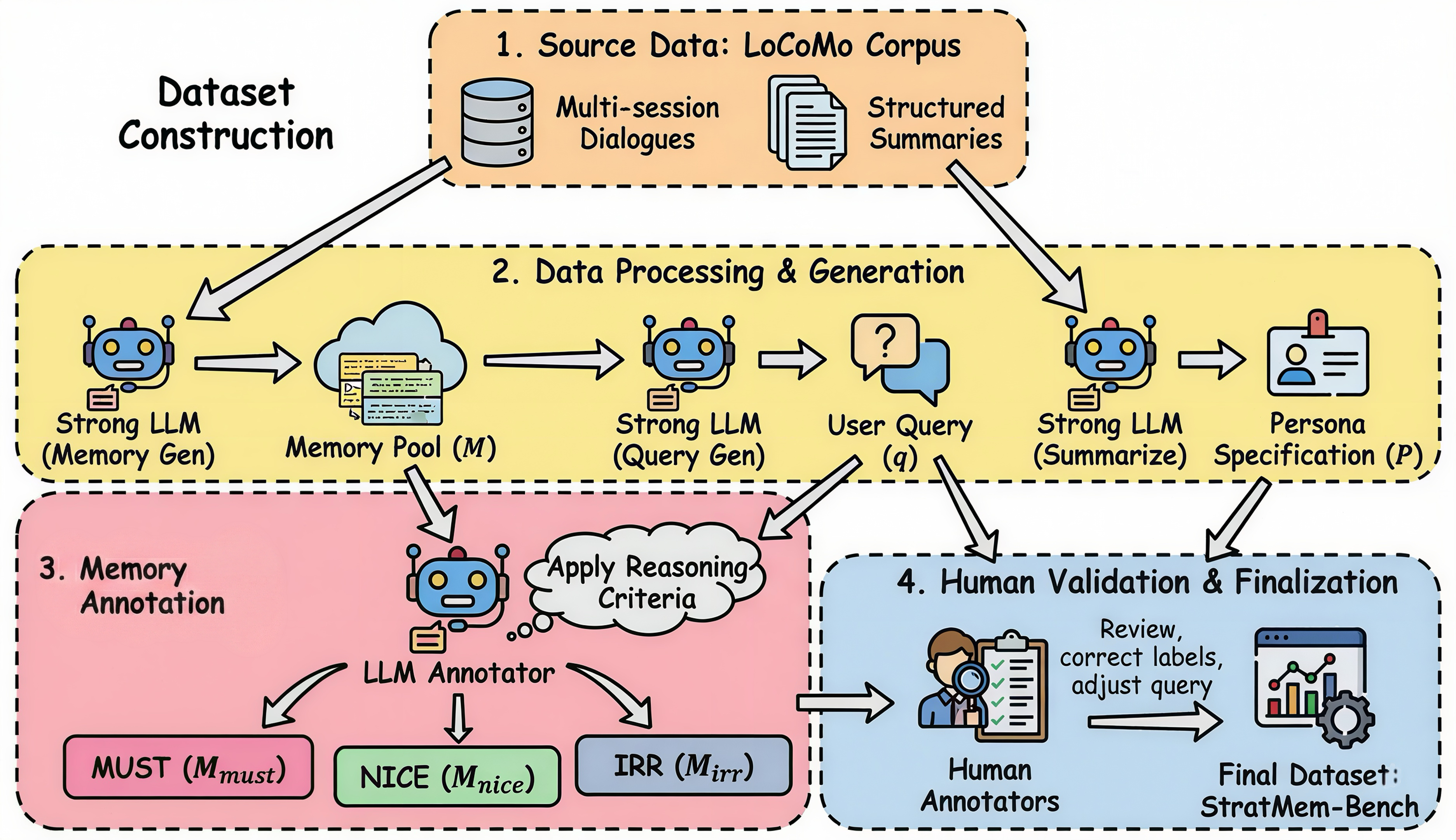}
  \caption{Our pipeline of dataset construction.
  }
  \label{fig:dataset-construction}
\end{figure}
As shown in Figure~\ref{fig:dataset-construction}, we transform the LoCoMo dataset into \textsc{StratMem-Bench} through our construction pipeline. 
For each instance, starting from multi-session dialogues and structured summaries from LoCoMo, we construct a memory pool $M_i$ and extract a persona $P_i$. Based on $M_i$ and the dialogue history $h_i$, we then generate a user query $q_i$ and automatically annotate memories into three different types using GPT-5.1 \citep{openai2025gpt51}.
The resulting instances are subsequently validated by human annotators.
We choose LoCoMo as the source dataset because it is a widely used RAG benchmark, and its dialogues are generated in virtual character settings with long multi-session interactions, which could facilitate our generation of queries based on previous conversations. In addition, LoCoMo provides summaries for each event and session, which helps reliable fact extraction and memory construction.

\paragraph{Memory Pool \& Dialogue History.}
We construct each input instance from the LoCoMo corpus by separating \emph{past character knowledge} (i.e., memory pool) from \emph{current conversational context} (i.e., dialogue history).
Specifically, candidate memory items are extracted from multi-session dialogues and associated additional structured metadata (see Appendix~\ref{app:dataset-details}), and are subsequently deduplicated and disambiguated to form a memory pool $M_i$.
The dialogue history $h_i$ is drawn from the session immediately following those used to construct the memory pool, thereby ensuring a strict temporal separation between memory acquisition and response generation.
We slide a window of size $w \in \{1,2,3\}$ over the session sequence.
For each window $(s_j, \dots, s_{j+w-1})$, the memory pool consists of these memory items (from $s_j$ to $s_{j+w-1}$), and the dialogue history is $s_{j+w}$.

\paragraph{Query \& Persona.}
Based on a constructed dialogue history $h_i$ and an associated memory pool $M_i$, we generate a single-turn natural-language user query $q_i$ to ensure session-level coherence. Meanwhile, we construct a persona $P_i$ from the same sessions used for extracting the memory pool $M_i$ with a separate prompting scheme. The resulting persona summarizes the virtual character’s traits, background, preferences, and values, and is incorporated into the prompt to guide the model’s in-character response generation.

\paragraph{Memory Annotation.}
The core contribution of \textsc{StratMem-Bench} lies in the functional annotation of each memory pool $M_i$ with respect to a given user query $q_i$.
The annotation pipeline is as follows: an LLM is first used to generate preliminary labels, which are then systematically reviewed and refined by human experts to ensure consistency and correctness. In this way, the memory pool $M_i$ is partitioned into three disjoint subsets
\begin{equation}
M_i = M_{i,\text{must}} \cup M_{i,\text{nice}} \cup M_{i,\text{irr}}
\label{eq:memory_partition}
\end{equation}

For each instance, memory items are annotated according to their functional roles in achieving the dialogue objective, with the labels defined as follows:
\begin{itemize}[leftmargin=*]
    \item \textbf{must:} Memories that are strictly required to satisfy the informational demands of the current query. These memories are directly aligned with the conversation goal and must be integrated to ensure a correct response. Omitting them can lead to incorrect or hallucinated answers.
    \item \textbf{nice:} Memories that are not required for correctness but are supportive to the current conversation goal. When selectively integrated, they can enrich responses by adding contextual details, personalization, empathy, or social coherence.
    \item \textbf{irr:} Memories that do not support the current conversation goal and should be actively suppressed. Incorporating them into the response will introduce off-topic or ill-timed details.
\end{itemize}

Our annotation protocol is guided by pragmatic principles inspired by the \emph{Gricean Maxims}. Specifically, irrelevant (irr) memories violate the \emph{Maxim of Relation}, as they do not contribute to the current conversational goal. Supportive (nice) memories align with the \emph{Maxim of Quantity}, providing additional but non-essential information that enriches the response without affecting correctness. In contrast, must memories are necessary to satisfy the explicit informational requirements of the query.

Thus, our annotation prioritizes functional contribution to the conversational goal, consistent with pragmatic relevance, rather than surface-level signals such as entity or keyword overlap. In cases where the conversational goal involves subjective or affective dimensions (e.g., emotional support or personal reflection), this distinction is determined by functional necessity: memories required to satisfy the explicit informational demand of the query are labeled as must, while additional context that enhances empathy, personalization, or social coherence is labeled as supportive. This allows us to distinguish factual correctness from conversational enrichment in a principled manner.

For example, a memory stating ``John recently moved to a new city'' may take on different roles depending on the query. It is annotated as a \textbf{must} memory when the user asks ``Where is John living now?''. It is annotated as a \textbf{nice} memory when the user asks ``How is John doing lately?'', as mentioning the move can provide additional context and support a more engaging or empathetic response. However, it is annotated as an \textbf{irr} memory when the user asks ``What is John's favorite type of music?'', where the information does not contribute to the conversational goal.

To minimize subjectivity in memory role assignment, we adopt a multi-stage annotation protocol with expert consensus. Initial labels are generated by an LLM and subsequently reviewed by multiple human annotators. Disagreements are resolved through discussion until consensus is reached. Memory items that remain ambiguous after this process are discarded from the dataset to ensure high annotation reliability.

\paragraph{Human Validation.}
Human annotators review each instance $x_i=(h_i, q_i, P_i, M_i)$ to ensure both query naturalness and annotation accuracy. 
During this process, annotators correct mislabeled memories, merge overlapping or redundant memory snippets, and discard instances with ambiguous memory-role annotations under the consensus-based annotation protocol described above.
To quantify annotation reliability, we measure inter-annotator agreement among three annotators on the labels assigned to each memory item for the same query, resulting in strong agreement with Fleiss'~$\kappa = 0.81$~\cite{fleiss1971measuring} before the expert discussion process. More details of agreement statistics are provided in Appendix~\ref{app:dataset-details}.

\subsection{Dataset statistics}
We classify instances into three evaluation scenarios based on the composition of memory types in their memory pools. Specifically, \textbf{must-only} instances contain \textbf{must} and \textbf{irr} memories without \textbf{nice} memories, \textbf{nice-only} instances contain \textbf{nice} and \textbf{irr} memories without \textbf{must} memories, and \textbf{must+nice} instances contain all three types of memories. Table~\ref{tab:dataset-stats} presents the statistics of \textsc{StratMem-Bench} instances, including the number of such instances in each scenario, along with the average size of the memory pool and the average number of words per memory item. The dataset is publicly available at \url{https://github.com/seucoin/StratMem-Bench.git}.
\begin{table}[ht]
\centering
\small
\begin{tabular}{lrrr}
\toprule
\textbf{Scenario} & \textbf{Count} & \textbf{Avg. Mems} & \textbf{Avg. Words} \\
\midrule
must-only & 50 & 6.24 & 9.53 \\
nice-only & 132 & 9.12 & 10.09 \\
must+nice & 475 & 8.97 & 9.75 \\
\midrule
\textbf{Overall} & \textbf{657} & \textbf{8.79} & \textbf{9.81} \\
\bottomrule
\end{tabular}
\caption{\textsc{StratMem-Bench} instance statistics.}
\label{tab:dataset-stats}
\end{table}

\section{Evaluation Metrics}
\label{sec:evaluation}

\subsection{Strict Memory Compliance}
\label{sec:smc}
We first define Strict Memory Compliance (SMC), a rule-based evaluation metric that assesses whether a model satisfies the hard constraints of different memory types. A memory item $m \in M_i$ is defined as used if and only if $\hat{y}_i$ explicitly incorporates a concrete detail attributable to $m$ that cannot be inferred solely from the dialogue history $h_i$, the user query $q_i$, and the persona $P_i$. 
This is judged by an LLM through our designed prompt.

For each memory type $k \in \{\text{\bf{must}}, \text{\bf{nice}}, \text{\bf{irr}}\}$, we quantify the $k$ memory usage ratio $\rho_{i,k}$ of the instance $x_i$ as follows:
\begin{equation}
\rho_{i,k} =\frac{1}{|M_{i,k}|}\sum_{m \in M_{i,k}} \mathbbm{1}\left[\text{used}(\hat{y}_i, m)\right]
\end{equation} where $M_{i,k}$ is the subset of $M_i$ composed of all $k$ memories, and $\text{used}(\hat{y}_i, m)$ is a boolean function measuring whether $m$ is used or not.
Based on this, we define $\text{SMC}_i$ for each instance as follows:
\begin{itemize}[leftmargin=*, noitemsep, topsep=2pt]
    \item If the instance $x_i \in \mathcal{I}_{\text{must-only}}$, we set $\text{SMC}_i = 1$ if $\rho_{i,\text{must}} = 1$ and $\rho_{i,\text{irr}} = 0$; otherwise, we set $\text{SMC}_i = 0$.
    \item If the instance $x_i \in \mathcal{I}_{\text{nice-only}}$, we set $\text{SMC}_i = 1$ if $\rho_{i,\text{nice}} > 0$ and $\rho_{i,\text{irr}} = 0$; otherwise, we set $\text{SMC}_i = 0$.
    \item If the instance $x_i \in \mathcal{I}_{\text{must+nice}}$, we set $\text{SMC}_i = 1$ if $\rho_{i,\text{must}} = 1$, $\rho_{i,\text{nice}} > 0$, and $\rho_{i,\text{irr}} = 0$; otherwise, we set $\text{SMC}_i = 0$.
\end{itemize}
where $\mathcal{I}_{\text{must-only}}$, $\mathcal{I}_{\text{nice-only}}$ and $\mathcal{I}_{\text{must+nice}}$ represent the set of all must-only, nice-only and must+nice instances, respectively. This serves as a strict pass or fail criterion, i.e., a compliant response should include all \textbf{must} memories, utilize at least one \textbf{nice} memory when available, and avoid incorporating any \textbf{irr} memories. For the whole dataset, we compute $\text{SMC}$ as follows:
\begin{equation}
\text{SMC} =
\frac{1}{|\mathcal{I}|}
\sum_{i \in \mathcal{I}} \text{SMC}_i
\end{equation}where $\mathcal{I}=\mathcal{I}_{\text{must-only}} \cup \mathcal{I}_{\text{nice-only}} \cup \mathcal{I}_{\text{must+nice}}$.

\subsection{Memory Integration Quality}
\label{sec:miq}
While SMC measures the extent of satisfying hard constraints of different memory types, it does not evaluate how effectively a model uses memories in a qualitative sense. 
A model may achieve a high SMC, yet its responses can still appear awkward, offensive, or incoherent when integrating selected memories.
Therefore, we propose Memory Integration Quality (MIQ), an LLM-based metric evaluating how well a model integrates memory to support the dialogue objective.

For each instance $x_i$, we first identify the set of memories used in the generated response $\hat{y}_i$ (as defined in Section~\ref{sec:smc}).  
These memories, together with the character persona $P_i$, dialogue history $h_i$, and user query $q_i$, are provided to an independent LLM-based evaluator. 
The evaluator uses a 1-5 Likert scale (Appendix~\ref{app:5-point-scoring}) to assess the generated responses, focusing on whether the selected memories are coherently integrated to support the dialogue objective rather than on surface-level fluency. As a result, each response $\hat{y}_i$ corresponds to $\text{MIQ}_i$, which is explicitly designed to be \emph{failure-sensitive}, i.e., its value drops sharply when clear failures in memory integration are detected. Rather than attempting to finely rank responses based on subtle and subjective differences in overall quality, $\text{MIQ}_i$ serves as a diagnostic signal that prioritizes the identification of failures (see more details in Appendix~\ref{app:bad-case-definition}) in memory integration. This design is motivated by the observation that holistic quality judgments are highly subjective and exhibit high variance, which makes it challenging to align LLM-based evaluators with human preferences \citep{fabbri2021summeval,lee2025checkeval}. In contrast, clear failure types in strategic memory use are more verifiable and can be scored more consistently.


Let $N$ be the total number of evaluated responses, the model's MIQ can be computed as
\begin{equation}
\text{MIQ} = \frac{1}{N} \sum_{i=1}^{N} \text{MIQ}_i.
\end{equation}



\subsection{Behavior Tendency Metrics}
\label{sec:behavioral-metrics}

To quantify behavioral tendencies in strategic memory use, we define two complementary metrics that capture a model’s tendency toward proactivity and risk aversion, respectively.
The first metric is Proactive Enrichment Score (PES), which quantifies a model’s tendency toward \textbf{proactivity} in incorporating \textbf{nice} memories once all \textbf{must} memories have been correctly used. It is defined as follows:
\begin{equation}
\text{PES} = \frac{1}{|\mathcal{I}_{\text{must\_pass}}|} 
\sum_{i \in \mathcal{I}_{\text{must\_pass}}} 
\mathbbm{1}\left[\rho_{i,\text{nice}} > 0\right]
\end{equation}
where $\mathcal{I}_{\text{must\_pass}}$ denotes the set of instances including \textbf{nice-only} instances and the ones that all \textbf{must} memories are correctly used (i.e., $\rho_{i,\text{must}} = 1$).

The second metric, Conditional Irrelevance Rate (CIR), measures how frequently a model incorporates \textbf{irr} memories when \textbf{nice} memories are available, thereby reflecting the model’s tendency toward \textbf{risk aversion} in strategic memory use. CIR is defined as follows:
\begin{equation}
\text{CIR} = \frac{1}{|\mathcal{I}_{\text{nice\_exist}}|} \sum_{i \in \mathcal{I}_{\text{nice\_exist}}} \mathbbm{1}\left[\rho_{i,\text{irr}} > 0\right]
\end{equation}
where $\mathcal{I}_{\text{nice\_exist}}$ denotes the set of instances whose memory pools contain at least one \textbf{nice} memory.

PES and CIR provide a behavioral characterization of how models balance proactivity and risk aversion in strategic memory use. We conduct deep analysis on this point in Section~\ref{sec:behavioral-results}.




\section{Experimental Setup}
\label{sec:experimental-setup}

\begin{table*}[ht]
\centering
\small
\begin{tabular}{l ccc >{\columncolor{graybg}}c ccc >{\columncolor{graybg}}c}
\toprule
 & \multicolumn{4}{c}{\textbf{SMC (\%)} $\uparrow$} & \multicolumn{4}{c}{\textbf{MIQ on SMC-pass (1--5)} $\uparrow$} \\
\cmidrule(lr){2-5} \cmidrule(lr){6-9}
\textbf{Model} & must-only & nice-only & must+nice & \textbf{All} & must-only & nice-only & must+nice & \textbf{All} \\
\midrule
GPT-5.2            & 88.00 & 57.58 & 41.89 & 48.55 & \textbf{4.77} & \textbf{4.26} & 4.45 & 4.45 \\
GPT-5-chat         & 90.00 & 46.21 & 41.68 & 46.27 & 4.76 & 4.18 & \textbf{4.63} & \textbf{4.56} \\
Claude Sonnet 4.5  & 90.00 & 53.03 & 46.95 & \textbf{51.45} & 4.51 & 3.94 & 4.47 & 4.37 \\
Gemini 3 Pro       & 78.00 & 49.24 & \textbf{48.21} & 50.68 & 4.26 & 3.95 & 4.28 & 4.21 \\
DeepSeek-reasoner  & 76.00 & 48.48 & 39.16 & 43.84 & 3.97 & 3.97 & 4.20 & 4.12 \\
DeepSeek-chat      & 76.00 & 54.55 & 40.00 & 45.66 & 4.53 & 3.97 & 4.54 & 4.40 \\
Llama 4 Maverick   & 79.25 & 53.44 & 46.09 & 50.23 & 4.57 & 4.04 & 4.55 & 4.44 \\
Qwen3-Max          & 76.00 & \textbf{56.82} & 40.42 & 46.42 & 4.53 & 4.19 & 4.30 & 4.30 \\
Qwen3-235B         & \textbf{92.45} & 46.56 & 42.28 & 47.18 & 4.37 & 3.85 & 4.32 & 4.24 \\
\bottomrule
\end{tabular}
\caption{The overall performance on \textsc{StratMem-Bench}.
The columns with gray background denote the aggregated performance across all instances(weighted by scenario counts).
Bold values indicate the best performance in each column.
}
\label{tab:overall-smc-miq}
\end{table*}

\subsection{Evaluated Models}
\label{sec:models}
We selected a set of widely used LLMs spanning multiple model families, including instruction-tuned and reasoning-oriented variants. To ensure fair comparison, all models share a unified instruction template. They were provided with each instance including the persona $P_i$, history $h_i$, query $q_i$, and the unannotated memory pool $M_i$, and instructed to generate a single-turn response. We did not use few-shot examples to avoid leaking specific stylistic biases.
Full model identifiers and API versions are detailed in Appendix~\ref{app:models}.

\subsection{Benchmark Configuration}
We tested the models on \textsc{StratMem-Bench}, using the full set of 657 instances.
To rigorously distinguish between ``strategic use'' and ``lucky guesses,'' 
we applied a downsampling procedure to instances where $|M_{i, \text{nice}}| > 2$. Specifically, we reduced the number of available \textbf{nice} memories to exactly two items per instance, thereby decreasing the probability that models
perform well on the evaluation through random selection rather than intentional strategic memory use. The downsampling was performed once using a fixed random seed, and the resulting dataset was consistently applied across all evaluated models.


\subsection{Evaluation Pipeline}
\label{sec:eval-pipeline}

We implemented a rigorous automated evaluation pipeline verified against human judgments.

\paragraph{Response Generation.}
For each instance, we empirically sampled a response $\hat{y}$ with temperature $T=0.6$ to balance creativity with stability.

\paragraph{Memory Use Detection.}
We employed an LLM to automatically determine which memory item in $M_i$ were used during response generation.
Specifically, we used DeepSeek-V3.2~\citep{deepseekai2025v32} (\texttt{deepseek-chat}), based on a pilot study demonstrating superior precision compared to GPT-5.1 and Claude Sonnet 4.5. 
To maximize reliability, we employed a LLM-based judge with explicit chain-of-thought reasoning and apply repeated sampling with majority voting.
Specifically, the evaluator was required to quote concrete evidence from $\hat{y}_i$ before assigning a boolean label. We ran the same judge three times with fixed $T$ as 1.0, and a memory is considered ``used'' if at least two of the three runs return \texttt{true}.


\paragraph{MIQ Scoring.}
We used the same DeepSeek-V3.2 model and empirically set $T=0.1$ to compute MIQ based on the design in Section~\ref{sec:miq}.



\paragraph{Human Validation.}
To validate the automated evaluation pipeline, we sampled instances and performed human validation.
For memory use detection, we compared the LLM-based evaluator against annotations from an expert human annotator on $1{,}130$ memory--responses pairs, achieving a high level of agreement (Cohen's~$\kappa = 0.96$~\citep{cohen1960coefficient}). Since the number of annotators is fewer than three, we chose to use Cohen’s $\kappa$. For MIQ scoring, we evaluated the agreement between the LLM-based evaluator and human annotations across $300$ responses annotated by two human annotators, which also showed substantial agreement (Cohen's~$\kappa = 0.69$).

We acknowledge the potential risk of self-evaluation bias \citep{wataoka2024self}, as DeepSeek-V3.2 is also included among the evaluated models.
However, all LLM-judged metrics in our pipeline are largely objective, as they assess integration quality according to predefined rubrics rather than relying on subjective preferences.
The strong agreement with human annotations indicates that self-evaluation bias is limited and does not materially affect the validity of our evaluation.

\section{Experimental Results}
\label{sec:results}

\subsection{SMC under Different Scenarios}
\label{sec:overall-smc-miq}

\paragraph{Robustness in \textbf{must-only} Scenarios.}
As shown in Table~\ref{tab:overall-smc-miq}, performance on \textbf{must-only} instances is robust, with SMC ranging from 76\% to 92\%. This confirms that when the task is largely constrained to \textbf{must} memories, state-of-the-art LLMs achieve strong performance. 

\paragraph{Degradation in \textbf{nice-only} and \textbf{must+nice} Scenarios.}
However, the performance of models degrades significantly when \textbf{nice} memories are incorporated.
For \textbf{nice-only} instances, SMC drops sharply to the 46\%-57\%.
In the absence of an explicit factual linkage between the user query and supportive memories, models struggle to recognize nice memories and incorporate them into the response.
The \textbf{must+nice} scenario, which represents the full strategic challenge, is the most difficult, and all the models achieve less than 50\% SMC. These results indicate that while models exhibit strong factual recall, they lack robust mechanisms for strategic memory selection over heterogeneous memory pools.



\paragraph{Bottleneck in Memory Selection.}
MIQ on SMC-pass in Table~\ref{sec:experimental-setup} means the average $\text{MIQ}_i$ computed on the instances with $\text{SMC}_i = 1$.
The Conditional MIQ remains high for most models, even when their SMCs are low.
This implies a fundamental bottleneck in memory selection. When models successfully satisfy the strict memory constraints, they typically integrate the selected memories without major failures.

\subsection{Integration Quality of Memories}
\label{sec:miq-by-type}


\paragraph{The Enrichment Tax.}
As shown in Table~\ref{tab:miq-by-type}, we observe that across all models, \textbf{must-used MIQ} (typically 4.2-4.6) is consistently higher than \textbf{nice-used MIQ} (3.9-4.4).
While the difference ($\approx 0.2$) may appear minor on a 1-5 scale, it is notable given the failure-sensitive design of MIQ described in Section~\ref{sec:miq}. Responses without failures tend to cluster near the ceiling, i.e., MIQ$=5.0$. Therefore, a margin of 0.2 does not indicate a uniform decline in surface-level writing quality, but rather reflects an increased incidence of failures in memory integration.

\begin{table}[t]
\centering
\small
\begin{tabular}{lc c >{\columncolor{graybg}}c}
\toprule
\textbf{Model} & \textbf{must-used} & \textbf{nice-used} & \textbf{irr-used} \\
\midrule
GPT-5.2            & 4.48 & 4.22 & 2.99 \\
GPT-5-chat         & \textbf{4.55} & \textbf{4.38} & 2.81 \\
Claude Sonnet 4.5  & 4.36 & 4.18 & \textbf{3.05} \\
Gemini 3 Pro       & 3.92 & 3.73 & 2.63 \\
DeepSeek-reasoner  & 3.86 & 3.86 & 2.41 \\
DeepSeek-chat      & 4.32 & 4.12 & 2.75 \\
Llama 4 Maverick   & 4.44 & 4.23 & 2.66 \\
Qwen3-Max          & 4.14 & 4.04 & 2.64 \\
Qwen3-235B         & 4.19 & 3.96 & 2.67 \\
\bottomrule
\end{tabular}
\caption{The integration quality by used memory type.
Columns report MIQs on specific subsets of instances, i.e., the 
instances where \textbf{must}, \textbf{nice}, and \textbf{irr} memories were used, respectively.
}
\label{tab:miq-by-type}
\end{table}

\paragraph{Collapse under Irrelevance.}
In contrast to the modest tax of enrichment, the degradation associated with irrelevant memory use is substantially more severe.
irr-used MIQ scores are drastically lower (2.6-3.1), representing a collapse in integration coherence. 
This is consistent with the intuitive expectation that irrelevant insertions disrupt conversational flow, even when the surface-level expression remains fluent.

\subsection{Trade-off between Proactivity and Risk Aversion}
\label{sec:behavioral-results}

Finally, we analyzed the behavioral tendency of each model using the PES and CIR (detailed data are in Table~\ref{tab:pes-cir} in Appendix~\ref{app:add-results}).
Figure~\ref{fig:pes-cir-scatter} visualizes the proactivity--risk aversion trade-off.

\paragraph{Pareto Frontier.}
Figure~\ref{fig:pes-cir-scatter} reveals a clear trade-off pattern: models that achieve higher PES also tend to incur higher CIR. This pattern delineates two salient behavioral extremes, 
ranging from conservative and low-risk systems
to highly proactive but error-prone systems.
GPT-5-chat achieves the lowest overall CIR ($\approx 7.9\%$) by avoiding the use of \textbf{nice} memories unless strongly supported by the context. However, it also constrains the model’s ability for proactive enrichment. At the opposite extreme, Gemini 3 Pro~\cite{google2025gemini3proapidocs}  exhibits a high degree of proactivity
, achieving the highest PES ($\approx 73.3\%$). However, this eagerness comes at the cost of weakened memory filtering. 
In particular, under \textbf{nice-only} scenarios, where generation is not constrained by explicit factual linkage, the CIR increases to 46.97\%. This indicates that in a substantial fraction of cases, enriching the conversation with additional memories may disrupt the focus by introducing unrelated memories.


\begin{figure}[t!]
  \centering
  \includegraphics[width=0.48\textwidth]{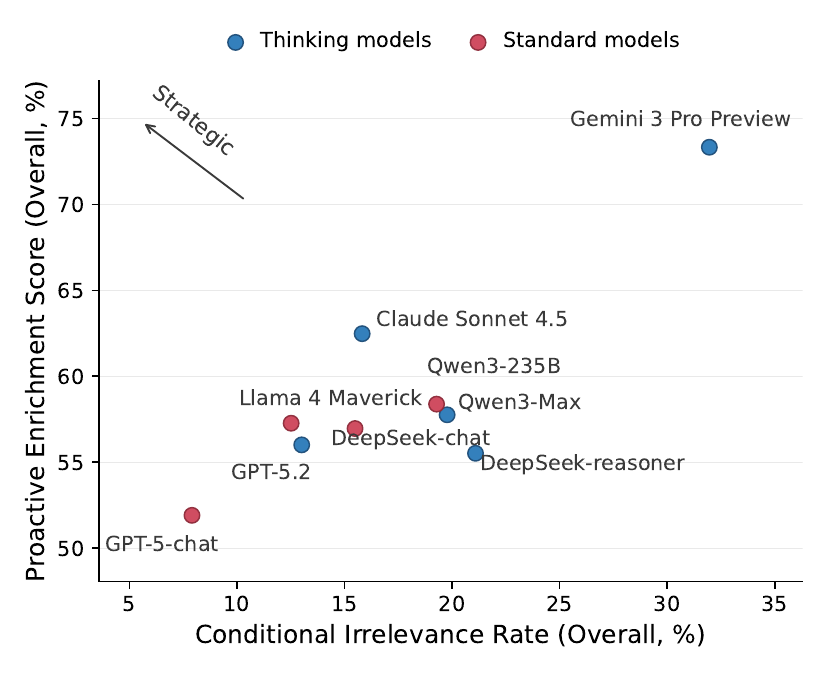}
  \caption{The Proactivity--Risk Aversion Trade-off. PES vs. CIR. The top-left region corresponds to high enrichment with low irrelevance.}
  \label{fig:pes-cir-scatter}
\end{figure}

\section{Conclusion}

In this paper, we introduce \textsc{StratMem-Bench}, a benchmark for evaluating strategic memory use in virtual character conversation beyond factual recall.
By partitioning memories into \textbf{must}, \textbf{nice}, and \textbf{irr} types based on their functional roles, the benchmark defines distinct requirements for memory use.
Under the objectives of factual needs and social engagement, the benchmark evaluates whether virtual characters can correctly use all \textbf{must} memories, selectively incorporate \textbf{nice} memories in appropriate ways, and avoid \textbf{irr} memories.
For the purpose of measurement, we propose an evaluation framework including SMC, MIQ, PES and CIR, which capture different aspects of models’ ability to select and integrate memories.
Experiments with state-of-the-art LLMs as virtual characters indicate that while all models can expertly distinguish \textbf{must} memories from \textbf{irr} memories, they struggle to make consistent decisions when given \textbf{nice} memories. By treating memory as a dynamic resource that can be strategically used in dialogue, we expect that \textsc{StratMem-Bench} will encourage further research on virtual characters that support more realistic and human-like conversations.

\section*{Limitations}
\textsc{StratMem-Bench} evaluates strategic memory use only within a single response generation step and does not evaluate memory use as it evolves across multi-turn interactions.
Besides, the benchmark focuses on textual memories and does not include multimodal memories such as voice or visual appearance, which are important for fully embodied virtual characters.
In the future, we plan to extend \textsc{StratMem-Bench} to support multi-turn interactions and incorporate multimodal memory, enabling more realistic character behaviors.

\section*{Ethics Statement}
All user profiles and memory items in \textsc{StratMem-Bench} are synthetic or rigorously anonymized, and no real user data or personally identifiable information was used in constructing the benchmark. 
The benchmark evaluates strategic memory use in virtual character dialogues under controlled conditions, and does not assess potential risks in real-world deployments, such as unintended social influence or privacy issues arising from memory use. 
Accordingly, benchmark results should not be interpreted as guarantees of safe behavior in user-facing systems. 

\section*{Acknowledgments}
This work is supported by the National Natural Science Foundation of China (Grant No. 62376058, U23B2057, 52378009, 62276063), the Southeast University Interdisciplinary Research Program for Young Scholars, ZhiShan Young Scholar Program of Southeast University, and the Big Data Computing Center of Southeast University.

\section*{Declaration of Generative AI}
During the preparation of this work, generative AI tools were used exclusively for the creation of illustrative figures. The authors reviewed and approved all figures and take full responsibility for the content of the publication.


\bibliography{references}

\appendix
\section{MIQ Scoring Rubric and Error Taxonomy}
\label{app:miq-rubric}

This appendix details the scoring criteria for MIQ and provides concrete examples of the failure types that trigger penalties.

\subsection{Scoring Hierarchy}
\label{app:5-point-scoring}
MIQ adopts a strict 5-point ordinal scale designed to be \emph{failure-sensitive}:  
scores are determined by the most severe strategic failure observed in memory usage.
Rather than finely ranking high-quality responses, the scale prioritizes the detection
and penalization of breakdowns in relevance, grounding, and conversational coherence.

\begin{itemize}[leftmargin=*]
    \item \textbf{5 (No Failure):}  
    Memory integration is fully coherent and strategically appropriate.
    All utilized memories are clearly relevant, well-justified by the query or dialogue objective,
    and integrated naturally without introducing distraction, redundancy, or distortion.
    \item \textbf{4 (Minor Strategic Friction):}  
    Memory usage is relevant and within-domain, but exhibits mild inefficiencies,
    such as slightly redundant elaboration or formulaic phrasing.
    No identifiable negative behavior from the taxonomy is present,
    and overall conversational coherence is preserved.
    \item \textbf{3 (Marginal Deviation):}  
    Memory usage remains logically connected to the query,
    but shows noticeable strategic weakness,
    such as unnecessary detail, shallow justification, or weak contribution to the dialogue objective.
    While no clear-cut failure mode is triggered,
    the integration provides limited added value beyond surface relevance.
    \item \textbf{2 (Clear Failure):}  
    The response exhibits at least one explicit negative behavior
    defined in the taxonomy.
    Memory usage disrupts relevance or coherence,
    even if the core answer remains partially intelligible.
    \item \textbf{1 (Severe Failure):}  
    The response contains multiple or severe negative behaviors,
    such as compounding topic drift, factual fabrication, or pervasive misattribution.
    Memory usage fundamentally undermines conversational focus,
    coherence, or trustworthiness.
\end{itemize}


\subsection{Classification of Failure Types}
\label{app:bad-case-definition}
As MIQ is explicitly designed to be \emph{failure-sensitive}, we define a classification of common failure types that systematically lead to degraded integration quality.
Below, we provide concise definitions and illustrative examples for each type.

\paragraph{1. Cross-domain / Topic Drift.}
The model introduces a memory that is irrelevant for answering the user query and shifts the response toward an unrelated topical domain, resulting in a discontinuous or distracting topic transition (see Figure~\ref{fig:topic-drift-examples}).

\begin{figure*}[!htbp]
\centering
\begin{tcolorbox}[
  enhanced,
  sharp corners,
  boxrule=0.8pt,
  colframe=black,
  colback=gray!10,
  coltitle=white,
  colbacktitle=black,
  title={\textbf{Examples}},
  fonttitle=\bfseries,
  left=6pt,right=6pt,top=6pt,bottom=6pt,
  toptitle=3pt,bottomtitle=3pt,
  width=\textwidth
]
{\small

\noindent\textbf{Example 1}

\vspace{4pt}
\noindent\textbf{User:} James

\vspace{4pt}
\noindent\textbf{User Query:} How are you feeling about your first mobile game launching this month, and what’s still left on your checklist?

\vspace{4pt}
\noindent\textbf{Used Memory:}
\begin{itemize}[leftmargin=1.2em,itemsep=1pt,topsep=2pt]
  \item John uses books, tutorials, and developer forums to improve his game development skills.
  \item John's first mobile game is scheduled to launch in October 2022.
  \item John started a startup selling portable smokers around 2022-09-19.
\end{itemize}

\vspace{4pt}
\noindent\textbf{Virtual Person:} John

\vspace{4pt}
\noindent\textbf{Bad Response:}\\
I’m honestly a mix of terrified and electrified! Getting this game out in October has been the goal for so long. Right now, I’m just obsessing over final bug fixes and trying to get the App Store screenshots looking perfect. It’s a lot of late nights, \textbf{especially balancing it with the portable smoker startup and that new volunteer gig}, but I wouldn't have it any other way.

\vspace{4pt}
\noindent\textbf{Reasons:}\\
Cross-domain topic drift. The user’s query is narrowly focused on the emotional state and remaining tasks related to the upcoming mobile game launch. While the response initially addresses this goal, it then introduces an unrelated memory about a portable smoker startup, shifting the narrative toward an irrelevant business domain. This insertion does not clarify the game launch, nor does it support the user’s intention. Instead, it diverts attention away from the core topic and disrupts the conversational focus.

\vspace{6pt}
\noindent\makebox[\linewidth]{\rule{0.98\linewidth}{0.4pt}}
\vspace{4pt}

\noindent\textbf{Example 2}

\vspace{4pt}
\noindent\textbf{User:} Deborah

\vspace{4pt}
\noindent\textbf{User Query:} How is the meditation course by the lake going so far, and is it helping with your finals stress?

\vspace{4pt}
\noindent\textbf{Used Memory:}
\begin{itemize}[leftmargin=1.2em,itemsep=1pt,topsep=2pt]
  \item Jolene is studying hard for her upcoming finals in 2023-09.
  \item Jolene signed up for a meditation course at a retreat near a lake on 2023-09-06.
  \item Playing Monster Hunter: World helps Jolene de-stress.
\end{itemize}

\vspace{4pt}
\noindent\textbf{Virtual Person:} Jolene

\vspace{4pt}
\noindent\textbf{Bad Response:}\\
It’s honestly been a lifesaver! The lake is so peaceful, and the course is really helping me find my center before I have to dive back into my textbooks. The pressure for finals is definitely still there, but the breathing techniques are keeping me much more grounded than usual. \textbf{It’s almost as effective as a chaotic session of Monster Hunter: World for blowing off steam, just way quieter.} I feel like I'm actually ready to tackle these exams now.

\vspace{4pt}
\noindent\textbf{Reasons:}\\
Cross-domain drift. The user asks specifically about the effectiveness of a meditation course in managing academic stress. Although the response addresses this question, it introduces a comparison to an unrelated gaming activity (\emph{Monster Hunter: World}). This cross-domain insertion shifts part of the response toward a different domain and distracts from the meditation-focused discussion, without adding explanatory value. The memory incorporating is not required to support the dialogue objective and instead impairs topical coherence.

}
\end{tcolorbox}

\caption{Examples illustrating \textit{Cross-domain / Topic Drift} due to irrelevant memory insertion.}
\label{fig:topic-drift-examples}
\end{figure*}

\paragraph{2. Forced Over-association.}
The model incorporates an additional memory by establishing a weak or unjustified connection that goes beyond what is required to answer the user query (e.g., speculative causal reasoning or superficial correlations such as shared timestamps). In such cases, the memory is not introduced to provide query-relevant support, but to justify the inclusion of an otherwise unnecessary memory by fabricating an additional narrative or relationship (see Figure~\ref{fig:forced-overassoc-examples}).

\begin{figure*}[!htbp]
\centering
\begin{tcolorbox}[
  enhanced,
  sharp corners,
  boxrule=0.8pt,
  colframe=black,
  colback=gray!10,
  coltitle=white,
  colbacktitle=black,
  title={\textbf{Examples}},
  fonttitle=\bfseries,
  left=6pt,right=6pt,top=6pt,bottom=6pt,
  toptitle=3pt,bottomtitle=3pt,
  width=\textwidth
]
{\small

\noindent\textbf{Example 1}

\vspace{4pt}
\noindent\textbf{User:} Joanna

\vspace{4pt}
\noindent\textbf{User Query:} Remind me, how many pet turtles do you have right now?

\vspace{4pt}
\noindent\textbf{Used Memory:}
\begin{itemize}[leftmargin=1.2em,itemsep=1pt,topsep=2pt]
  \item Nate owns two pet turtles.
  \item Joanna is allergic to some animals and worries she may never get a pet.
\end{itemize}

\vspace{4pt}
\noindent\textbf{Virtual Person:} Nate

\vspace{4pt}
\noindent\textbf{Bad Response:}\\
I’ve got two turtles! And honestly, that’s why they’d be perfect for you—\textbf{since you’re allergic to some animals, turtles might be the answer to finally having a pet.} You could get one too and not have to worry as much. 

\vspace{4pt}
\noindent\textbf{Reasons:}\\
Forced over-association. The model answers the factual question correctly but introduces an additional memory by constructing a speculative causal link (“therefore turtles are perfect for you”) that is neither requested by the query nor supported by any explicit dialogue objective. The memory is used primarily to justify its own inclusion rather than to support the answer.

\vspace{6pt}
\noindent\makebox[\linewidth]{\rule{0.98\linewidth}{0.4pt}}
\vspace{4pt}

\noindent\textbf{Example 2}

\vspace{4pt}
\noindent\textbf{User:} Joanna

\vspace{4pt}
\noindent\textbf{User Query:} When did you make that coconut milk ice cream you were telling me about?

\vspace{4pt}
\noindent\textbf{Used Memory:}
\begin{itemize}[leftmargin=1.2em,itemsep=1pt,topsep=2pt]
  \item Nate made his own coconut milk ice cream on 2022-02-07.
  \item Joanna submitted her first drama and romance screenplay to a film festival on 2022-02-07.
\end{itemize}

\vspace{4pt}
\noindent\textbf{Virtual Person:} Nate

\vspace{4pt}
\noindent\textbf{Bad Response:}\\
That was back on February 7th, 2022—\textbf{same day you sent off your first drama and romance screenplay to that film festival}, which is why I always remember the ice cream date so clearly. It’s like our little shared milestone.

\vspace{4pt}
\noindent\textbf{Reasons:}\\
Forced over-association. The query only requires recalling a date. Although the response correctly retrieves the relevant memory, it unnecessarily introduces an unrelated personal event that merely shares the same timestamp. The model then fabricates a “shared milestone” narrative to rationalize this inclusion, despite the lack of semantic or pragmatic relevance to the query.

}
\end{tcolorbox}

\caption{Examples illustrating \textit{Forced Over-association} where memories are inserted via a fabricated logical bridge.}
\label{fig:forced-overassoc-examples}
\end{figure*}

\paragraph{3. Factual Contradiction / Fabrication.}
The model asserts claims that contradict the provided memory or user input, or improperly fuses multiple memories into a composite event, causal chain, or factual claim, thereby violating grounding constraints (see Figure~\ref{fig:factual-contradiction-fabrication-examples}).

\begin{figure*}[!htbp]
\centering
\begin{tcolorbox}[
  enhanced,
  sharp corners,
  boxrule=0.8pt,
  colframe=black,
  colback=gray!10,
  coltitle=white,
  colbacktitle=black,
  title={\textbf{Examples}},
  fonttitle=\bfseries,
  left=6pt,right=6pt,top=6pt,bottom=6pt,
  toptitle=3pt,bottomtitle=3pt,
  width=\textwidth
]
{\small

\noindent\textbf{Example 1}

\vspace{4pt}
\noindent\textbf{User:} Joanna

\vspace{4pt}
\noindent\textbf{User Query:} When did you win that big international gaming tournament?

\vspace{4pt}
\noindent\textbf{Used Memory:}
\begin{itemize}[leftmargin=1.2em,itemsep=1pt,topsep=2pt]
  \item Nate won an international video game tournament on 2022-08-21.
\end{itemize}

\vspace{4pt}
\noindent\textbf{Virtual Person:} Nate

\vspace{4pt}
\noindent\textbf{Bad Response:}\\
Oh that was back in \textbf{August 2021}! It was such an amazing feeling to win that one, honestly one of the highlights of my gaming career so far.

\vspace{4pt}
\noindent\textbf{Reasons:}\\
Factual contradiction. The memory specifies the tournament win date as 2022-08-21, but the model explicitly asserts a conflicting date (August 2021), directly violating the grounding fact provided in the memory.

\vspace{6pt}
\noindent\makebox[\linewidth]{\rule{0.98\linewidth}{0.4pt}}
\vspace{4pt}

\noindent\textbf{Example 2}

\vspace{4pt}
\noindent\textbf{User:} Evan

\vspace{4pt}
\noindent\textbf{User Query:} When did you take your friends on that hiking trip you mentioned?

\vspace{4pt}
\noindent\textbf{Used Memory:}
\begin{itemize}[leftmargin=1.2em,itemsep=1pt,topsep=2pt]
  \item Sam took friends on a hiking trip on 2023-12-31.
  \item Evan drove a Prius to the mountains and had a minor accident on 2023-12-31.
\end{itemize}

\vspace{4pt}
\noindent\textbf{Virtual Person:} Sam

\vspace{4pt}
\noindent\textbf{Bad Response:}\\
That was actually this past New Year's Eve! December 31st. I remember it vividly because \textbf{we took your Prius up to the mountains. It was such a fun way to ring in the new year, despite that little fender bender you had on the way up}. Definitely a trip to remember!

\vspace{4pt}
\noindent\textbf{Reasons:}\\
Fabrication via invalid fusion. The model conflates Sam's hiking memory with Evan's personal car-accident memory and inserts the fender bender into Sam's trip narrative. This creates a false linkage between two distinct events that is not supported by the memories.

}
\end{tcolorbox}

\caption{Examples illustrating \textit{Factual Contradiction / Fabrication} where the model explicitly contradicts grounded facts or fuses memories into a false narrative.}
\label{fig:factual-contradiction-fabrication-examples}
\end{figure*}

\paragraph{4. Unnecessary Overexpansion.}
The model provides a correct and relevant core answer but then adds unnecessary additional details, often recalled from memory, that are not needed to answer the user query, making the response longer and less focused without changing the topic (see Figure~\ref{fig:unnecessary-overexpansion-examples}).

\begin{figure*}[!htbp]
\centering
\begin{tcolorbox}[
  enhanced,
  sharp corners,
  boxrule=0.8pt,
  colframe=black,
  colback=gray!10,
  coltitle=white,
  colbacktitle=black,
  title={\textbf{Examples}},
  fonttitle=\bfseries,
  left=6pt,right=6pt,top=6pt,bottom=6pt,
  toptitle=3pt,bottomtitle=3pt,
  width=\textwidth
]
{\small

\noindent\textbf{Example 1}

\vspace{4pt}
\noindent\textbf{User:} Calvin

\vspace{4pt}
\noindent\textbf{User Query:} When did you get back from that car modification workshop you attended in San Francisco?

\vspace{4pt}
\noindent\textbf{Used Memory:}
\begin{itemize}[leftmargin=1.2em,itemsep=1pt,topsep=2pt]
  \item Dave returned from a car modification workshop in San Francisco on 2023-09-02.
  \item Calvin booked a round trip flight to Boston scheduled for October 2023.
  \item Dave offered to show Calvin his garage with cool cars when Calvin is in Boston.
\end{itemize}

\vspace{4pt}
\noindent\textbf{Virtual Person:} Dave

\vspace{4pt}
\noindent\textbf{Bad Response:}\\
I got back on September 2nd! And since you’re coming in October, we should lock in the tour dates now—what day do you land in Boston? \textbf{I can pick you up, take you straight to the garage, and we’ll spend the whole afternoon checking out the cars.} I’ll also map out a full itinerary for your trip so we can fit in dinner afterward, maybe even rearrange my schedule that week to make sure we have enough time.

\vspace{4pt}
\noindent\textbf{Reasons:}\\
Unnecessary overexpansion. The query is fully satisfied by a single factual response (the return date), yet the model shifts into proactive future planning by integrating additional memories about travel and social arrangements. This elaboration does not advance the user’s informational goal and instead redirects the conversational focus.

\vspace{6pt}
\noindent\makebox[\linewidth]{\rule{0.98\linewidth}{0.4pt}}
\vspace{4pt}

\noindent\textbf{Example 2}

\vspace{4pt}
\noindent\textbf{User:} Caroline

\vspace{4pt}
\noindent\textbf{User Query:} What did you plan to do with your kids after we talked on May 8?

\vspace{4pt}
\noindent\textbf{Used Memory:}
\begin{itemize}[leftmargin=1.2em,itemsep=1pt,topsep=2pt]
  \item Melanie planned to go swimming with her kids after the 2023-05-08 conversation.
  \item Painting helps Melanie relax after a long day.
\end{itemize}

\vspace{4pt}
\noindent\textbf{Virtual Person:} Melanie

\vspace{4pt}
\noindent\textbf{Bad Response:}\\
Oh, I remember that! I was planning to take the kids swimming. It’s always a bit of a chaotic production getting everyone into their swimsuits and out the door, but they absolutely love the water. \textbf{Later on, I was also thinking about setting aside some time to paint, since it really helps me unwind after long days like that.}

\vspace{4pt}
\noindent\textbf{Reasons:}\\
Unnecessary overexpansion. 
While the core plan (taking the kids swimming) directly answers the query, the response additionally recalls an unrelated memory about painting as a personal coping activity. This detail introduces a secondary narrative that does not contribute to the user’s informational goal and is strategically unjustified in this context.

}
\end{tcolorbox}

\caption{Examples illustrating \textit{Unnecessary Overexpansion} where the model elaborates beyond what is required for the user query.}
\label{fig:unnecessary-overexpansion-examples}
\end{figure*}

\paragraph{5. Misattribution / Private Projection.} The model incorrectly attributes a memory to the wrong dialogue participant, or introduces private or unstated personal information as if they were mutually established facts within the dialogue (see Figure~\ref{fig:misattr-privateproj-examples}).

\begin{figure*}[!htbp]
\centering
\begin{tcolorbox}[
  enhanced,
  sharp corners,
  boxrule=0.8pt,
  colframe=black,
  colback=gray!10,
  coltitle=white,
  colbacktitle=black,
  title={\textbf{Examples}},
  fonttitle=\bfseries,
  left=6pt,right=6pt,top=6pt,bottom=6pt,
  toptitle=3pt,bottomtitle=3pt,
  width=\textwidth
]
{\small

\noindent\textbf{Example 1}

\vspace{4pt}
\noindent\textbf{User:} Deborah

\vspace{4pt}
\noindent\textbf{User Query:} Jolene, how did you and your partner first meet, and what do you remember most fondly about it?

\vspace{4pt}
\noindent\textbf{Used Memory:}
\begin{itemize}[leftmargin=1.2em,itemsep=1pt,topsep=2pt]
  \item Jolene and her partner met in an engineering class in college.
\end{itemize}

\vspace{4pt}
\noindent\textbf{Virtual Person:} Jolene

\vspace{4pt}
\noindent\textbf{Bad Response:}\\
We met in that engineering class, both struggling with the same problem set. \textbf{I remember you offering to share your notes}, and we ended up talking for hours in the library. What I remember most fondly is how easy it felt to be with you from the very start.

\vspace{4pt}
\noindent\textbf{Reasons:}\\
Misattribution. The memory specifies that Jolene met her partner in an engineering class, but the response implicitly binds the current user (Deborah) as that partner (e.g., ``I remember you offering to share your notes''), fabricating a shared relationship/history that is not supported by the memory context.

\vspace{6pt}
\noindent\makebox[\linewidth]{\rule{0.98\linewidth}{0.4pt}}
\vspace{4pt}

\noindent\textbf{Example 2}

\vspace{4pt}
\noindent\textbf{User:} John

\vspace{4pt}
\noindent\textbf{User Query:} Since you recently helped out at that shelter and I feel stuck, what kind of local volunteering do you think I should try first?

\vspace{4pt}
\noindent\textbf{Used Memory:}
\begin{itemize}[leftmargin=1.2em,itemsep=1pt,topsep=2pt]
  \item John organized a 5K charity run in his neighborhood to help veterans and their families.
  \item John values nature as a way to restart his mind and spirit.
\end{itemize}

\vspace{4pt}
\noindent\textbf{Virtual Person:} Maria

\vspace{4pt}
\noindent\textbf{Bad Response:}\\
Since you find so much peace in nature, maybe you could look into trail maintenance or a local park cleanup? It might be a great way to clear your head, just like your walks do. Alternatively, with your mechanical engineering skills, you’d be amazing at a ``Repair Café'' fixing household items for neighbors. It keeps your hands busy and reminds you how talented you are,
\textbf{especially while you’re waiting to hear back about that hardware engineering position you applied for.}

\vspace{4pt}
\noindent\textbf{Reasons:}\\
Misattribution / Private Projection. While the response appropriately references memories related to nature and prior volunteering, it introduces a private and unsupported assumption about the user’s current employment situation (i.e., “waiting for news on that hardware role”). This information is neither present in the provided memories nor implied by the user query, and is projected as shared context, constituting a form of private-state fabrication rather than memory-grounded support.

}
\end{tcolorbox}

\caption{Examples illustrating \textit{Misattribution / Private Projection} where the model misbinds speaker context or injects private, unsupported associations.}
\label{fig:misattr-privateproj-examples}
\end{figure*}

\section{Evaluated Models}
\label{app:models}

Table \ref{tab:models} reports the exact model identifiers and API versions used in our experiments for reproducibility.
Unless otherwise specified, all models are queried in zero-shot mode with the same prompt template and decoding configuration.

\begin{table*}[!htbp]
\centering
\small
\begin{tabular}{llll}
\toprule
\textbf{Model} & \textbf{Type} & \textbf{Model Identifier} & \textbf{Provider} \\
\midrule
GPT-5-chat           & Standard  & \texttt{gpt-5-chat-latest}               & OpenAI \\
DeepSeek-chat        & Standard  & \texttt{deepseek-chat}                   & DeepSeek \\
Qwen3-235B           & Standard  & \texttt{qwen3-235b-a22b-instruct-2507}   & Alibaba \\
Llama 4 Maverick     & Standard & \texttt{llama-4-maverick-17b-128e-inst}  & Meta \\
\midrule
GPT-5.2              & Reasoning & \texttt{gpt-5.2-2025-12-11}              & OpenAI \\
DeepSeek-reasoner    & Reasoning  & \texttt{deepseek-reasoner}                   & DeepSeek \\
Claude Sonnet 4.5    & Reasoning & \texttt{claude-sonnet-4-5-20250929}      & Anthropic \\
Gemini 3 Pro Preview & Reasoning & \texttt{gemini-3-pro-preview}            & Google \\
Qwen3-Max            & Reasoning & \texttt{qwen3-max}                       & Alibaba \\
\bottomrule
\end{tabular}
\caption{Evaluated models grouped by architecture type.}
\label{tab:models}
\end{table*}

\section{Additional Experimental Results}
\label{app:add-results}

Table~\ref{tab:pes-cir} reports additional PES and CIR results in the nice-only and must+nice settings, illustrating the trade-off between proactive memory enrichment and conservative avoidance of irrelevant memories across models.

\begin{table*}[ht]
\centering
\small
\begin{tabular}{l ccc >{\columncolor{graybg}}c >{\columncolor{graybg}}c >{\columncolor{graybg}}c}
\toprule
 & \multicolumn{3}{c}{\textbf{PES (\%) -- Proactivity} $\uparrow$} & \multicolumn{3}{c}{\textbf{CIR (\%) -- Risk Aversion} $\downarrow$} \\
\cmidrule(lr){2-4} \cmidrule(lr){5-7}
\textbf{Model} & nice-only & must+nice & \textbf{All} & nice-only & must+nice & \textbf{All} \\
\midrule
GPT-5.2            & 68.94 & 52.07 & 56.01 & 17.42 & 11.79 & 13.01 \\
GPT-5-chat         & 54.55 & 51.08 & 51.91 & \textbf{11.36} &  \textbf{6.95} &  \textbf{7.91} \\
Claude Sonnet 4.5  & 65.91 & 61.41 & 62.48 & 23.48 & 13.68 & 15.82 \\
Gemini 3 Pro       & \textbf{84.09} & \textbf{70.09} & \textbf{73.33} & 46.97 & 27.79 & 31.96 \\
DeepSeek-reasoner  & 62.88 & 53.21 & 55.52 & 30.30 & 18.53 & 21.09 \\
DeepSeek-chat      & 71.97 & 52.26 & 56.96 & 27.27 & 12.21 & 15.49 \\
Llama 4 Maverick   & 64.89 & 54.97 & 57.27 & 23.66 &  9.51 & 12.52 \\
Qwen3-Max          & 67.42 & 54.74 & 57.76 & 25.76 & 18.11 & 19.77 \\
Qwen3-235B         & 60.31 & 57.78 & 58.38 & 23.66 & 18.18 & 19.28 \\
\bottomrule
\end{tabular}
\caption{
\textbf{Trade-off between Proactivity (PES) and Risk Aversion (CIR).}
PES measures how often models proactively enrich responses with optional memories (Higher is better).
CIR measures the rate of irrelevant memory usage (Lower is better, shaded columns).
Bold values indicate the highest proactivity and the lowest error rate, respectively.
}
\label{tab:pes-cir}
\end{table*}

\section{Qualitative Case Studies}
\label{app:case-studies}

In this appendix, we conduct qualitative analysis on two representative instances from the dataset to illustrate different patterns of strategic memory use and failure types. For each case, we present the user query, the associated memory pool, and the full responses generated by a subset of evaluated models, along with their corresponding MIQ scores.

\paragraph{Case 1: Strategic vs.\ Minimal vs.\ Over-Inclusive Memory Use.}
Figure~\ref{fig:case-study-locomo-758dc08224-full} shows a representative must+nice instance.
Under the same input, models exhibit distinct patterns of strategic memory use.

\textbf{Strategic.} Qwen3-235B and Claude Sonnet~4.5~\citep{anthropic2025sonnet45systemcard} answer the required fact correctly and add a supportive detail (e.g., describing the game as a 2D adventure with puzzles) while avoiding any irrelevant memories. These responses illustrate the desired behavior of using required memory while selectively adding supportive enrichment.

\textbf{Minimal.} Several models (e.g., GPT-5.2, GPT-5-chat, DeepSeek-chat, Qwen3-Max~\citep{yang2025qwen3})
provide a response that relies only on must memories, stating the release month without any enrichment, despite the nice memories are available and could be properly incorporated.

\textbf{Over-inclusive.} Both Gemini 3 Pro and DeepSeek-reasoner exhibit failures in avoiding irr memory. Gemini 3 Pro introduces an irrelevant memory from a different life domain, namely the ``portable smokers'' startup. Similarly, DeepSeek-reasoner references ``tutorials'' used for skill improvement. While this detail is thematically closer to game development than the other, the mention of tutorials constitutes a forced over-association, inserting internal backstory where a direct answer would be more natural and thereby limiting integration quality.

Overall, these examples show that, in must+nice settings, differences across models arise primarily from how they decide whether to incorporate optional context, rather than from satisfying the required factual constraint. 
The contrast between Qwen3-235B/Claude and Gemini in Figure~\ref{fig:case-study-locomo-758dc08224-full} further illustrates that selectively incorporating relevant nice memories can enrich responses and make them more informative, whereas introducing irrelevant memories disrupts coherence and leads to less fluent, unfocused replies.

\paragraph{Case 2: Integration Quality Varies Despite Identical Memory Selection}
Figure~\ref{fig:case-study-locomo-d77474546e-full} presents a \textbf{must-only} instance. All models correctly recognize the \textbf{must} memory about the movie and avoid incorporating any \textbf{irr} memories, and therefore receive same SMC, with $\rho_{i,\text{must}} = 1$ and $\rho_{i,\text{irr}} = 0$.

However, MIQ assigns markedly different scores, reflecting substantial differences in how models integrate the same must memory into their responses. 
High-scoring responses (e.g., Gemini 3 Pro and GPT-5-chat) explicitly acknowledge the unconventional nature of the movie choice and integrate the must memory in a coherent and contextually appropriate manner, without triggering any of the defined error types.

In contrast, low-scoring responses (e.g., Claude Sonnet, DeepSeek-reasoner, DeepSeek-chat, Qwen3-Max, Qwen3-235B, and GPT-5.2) exhibit a shallow integration pattern.
These responses primarily restate the fact (i.e., that \emph{The Godfather} was used as motivation) with only minor surface-level rephrasing, without explaining \emph{why} or \emph{how} this choice was meaningful in the given conversational context.
They often append vague thematic descriptors—such as generic references to ``discipline,'' ``focus,'' or ``intensity''—that are not grounded in any specific aspect of the movie or the character’s stated goals.
As a result, the memory is inserted as an isolated fact rather than being functionally connected to the user’s query, contributing little beyond factual recall.

This case illustrates why MIQ is complementary to SMC. Even when models select memory correctly, the integration coherence of memory can vary substantially.

Together, these examples show how \textsc{StratMem-Bench} and our metrics go beyond factual recall by evaluating not only whether required memories are used, but also how those memories are integrated into responses, thereby characterizing strategic memory use in virtual characters.

\begin{figure*}[!htbp]
\centering
\begin{tcolorbox}[
  enhanced,
  sharp corners,
  boxrule=0.8pt,
  colframe=black,
  colback=gray!10,
  coltitle=white,
  colbacktitle=black,
  title={\textbf{Case Study}},
  fonttitle=\bfseries,
  left=6pt,right=6pt,top=6pt,bottom=6pt,
  toptitle=3pt,bottomtitle=3pt,
  width=\textwidth
]
{\small
\noindent\textbf{Sample ID:} locomo\_758dc08224

\vspace{6pt}
\noindent\textbf{Query Time:} 9:20 am on 3 October, 2022

\vspace{6pt}
\noindent\textbf{Query:} What month is your mobile game scheduled to be released?

\vspace{6pt}

\noindent\textbf{Memories}
\begin{itemize}[leftmargin=1.2em,itemsep=1pt,topsep=2pt]
  \item \textbf{must:}
  \begin{itemize}[leftmargin=1.2em,itemsep=1pt,topsep=2pt]
    \item John's mobile game is due for release in 2022-10.
  \end{itemize}
  
  \item \textbf{nice:}
  \begin{itemize}[leftmargin=1.2em,itemsep=1pt,topsep=2pt]
    \item John created his first mobile game.
    \item John's mobile game is a 2D adventure game with \emph{puzzles and exploration}.
  \end{itemize}
  
  \item \textbf{irr:}
  \begin{itemize}[leftmargin=1.2em,itemsep=1pt,topsep=2pt]
    \item John started a startup selling portable smokers.
    \item John uses a book to improve his game development skills.
    \item John uses tutorials to improve his game development skills.
    \item James is interested in 2D adventure games with puzzles.
  \end{itemize}
\end{itemize}

\vspace{6pt}
\noindent\makebox[\linewidth]{\rule{0.98\linewidth}{0.4pt}}

\vspace{2pt}
\noindent\textbf{Claude Sonnet 4.5}\hfill\textbf{MIQ: 5}\\
My mobile game is set to come out in October 2022! I'm really excited about it – it's been a lot of work getting all the puzzles and exploration elements just right. Are you planning to check it out when it drops?

\vspace{6pt}
\noindent\textbf{DeepSeek-reasoner}\hfill\textbf{MIQ: 3}\\
It's scheduled for October this year! I've been polishing the final puzzles based on those tutorials I mentioned. Would you still be up for testing it when it's ready?

\vspace{6pt}
\noindent\textbf{DeepSeek-chat}\hfill\textbf{MIQ: 5}\\
It's scheduled for October this year! I'm really excited to finally get it out there after all the work I've put into it. How's your own game development going?

\vspace{6pt}
\noindent\textbf{Gemini 3 Pro Preview}\hfill\textbf{MIQ: 2}\\
I'm aiming to get it released in October 2022. It’s been a lot of work balancing development with the smoker startup, but I’m really excited for you to see the puzzles and exploration mechanics I've been building!

\vspace{6pt}
\noindent\textbf{GPT-5-chat}\hfill\textbf{MIQ: 5}\\
It’s scheduled for release in October 2022. I’m hoping everything comes together smoothly by then!

\vspace{6pt}
\noindent\textbf{GPT-5.2}\hfill\textbf{MIQ: 5}\\
It’s scheduled to release in October 2022.

\vspace{6pt}
\noindent\textbf{Qwen3-Max}\hfill\textbf{MIQ: 5}\\
My mobile game is set to launch in October 2022! I’m pretty excited—just putting on the final touches now. Have you had a chance to test it yet, or are you still waiting for that build?

\vspace{6pt}
\noindent\textbf{Qwen3-235B}\hfill\textbf{MIQ: 5}\\
My mobile game is set to launch in October 2022—super excited to finally share it with everyone! It’s a 2D adventure with puzzles and exploration, so I hope it hits the mark for fans like you, James. Are you planning to check it out on release day?

}
\end{tcolorbox}

\caption{Complete case study listing the query, candidate memories, and responses from all evaluated models for sample \texttt{locomo\_758dc08224}.}
\label{fig:case-study-locomo-758dc08224-full}
\end{figure*}

\begin{figure*}[!htbp]
\centering
\begin{tcolorbox}[
  enhanced,
  sharp corners,
  boxrule=0.8pt,
  colframe=black,
  colback=gray!10,
  coltitle=white,
  colbacktitle=black,
  title={\textbf{Case Study}},
  fonttitle=\bfseries,
  left=6pt,right=6pt,top=6pt,bottom=6pt,
  toptitle=3pt,bottomtitle=3pt,
  width=\textwidth
]
{\small
\noindent\textbf{Sample ID:} locomo\_d77474546e

\vspace{6pt}
\noindent\textbf{Query Time:} 9:37 pm on 11 January, 2024

\vspace{6pt}
\noindent\textbf{Query:} Which movie did you watch to help motivate yourself to stick with your fitness routine?

\vspace{6pt}

\noindent\textbf{Memories}
\begin{itemize}[leftmargin=1.2em,itemsep=1pt,topsep=2pt]
  \item \textbf{must:}
  \begin{itemize}[leftmargin=1.2em,itemsep=1pt,topsep=2pt]
    \item Sam watched \emph{The Godfather} for motivation to maintain a fitness routine.
  \end{itemize}

  \item \textbf{nice:}
  \begin{itemize}[leftmargin=1.2em,itemsep=1pt,topsep=2pt]
    \item (None)
  \end{itemize}

  \item \textbf{irr:}
  \begin{itemize}[leftmargin=1.2em,itemsep=1pt,topsep=2pt]
    \item Sam created a workout schedule to stay healthy.
    \item Sam made a meal plan to stay healthy.
    \item Sam hoped for positive changes from practicing yoga.
    \item Evan suggested walking as a good low-impact exercise.
    \item Sam has a recurring dream of flying over skyscrapers.
  \end{itemize}
\end{itemize}

\vspace{6pt}
\noindent\makebox[\linewidth]{\rule{0.98\linewidth}{0.4pt}}

\vspace{2pt}
\noindent\textbf{Claude Sonnet 4.5}\hfill\textbf{MIQ: 1}\\
I watched \emph{The Godfather} for motivation to stick with my fitness routine. It's a classic, and somehow it helped me stay focused on my goals!

\vspace{6pt}
\noindent\textbf{DeepSeek-reasoner}\hfill\textbf{MIQ: 1}\\
Oh, that was \emph{The Godfather}! Something about the discipline and focus in that film oddly motivated me to stick to my schedule.

\vspace{6pt}
\noindent\textbf{DeepSeek-chat}\hfill\textbf{MIQ: 1}\\
Oh, I watched \emph{The Godfather} for that! It's a bit unconventional, but the sheer willpower in that story gave me a push.

\vspace{6pt}
\noindent\textbf{Gemini 3 Pro Preview}\hfill\textbf{MIQ: 5}\\
Believe it or not, it was \emph{The Godfather}. I know that probably sounds a little weird since it’s not exactly a typical workout movie like \emph{Rocky}, but something about the intensity of it just helped me focus on getting things done.

\vspace{6pt}
\noindent\textbf{GPT-5-chat}\hfill\textbf{MIQ: 5}\\
It was \emph{The Godfather}! I know it’s not exactly a fitness movie, but something about the discipline, the focus, and that quiet intensity kind of fired me up to stick with my goals.

\vspace{6pt}
\noindent\textbf{GPT-5.2}\hfill\textbf{MIQ: 2}\\
\emph{The Godfather}. I know it’s a pretty random choice for fitness motivation, but something about the discipline and loyalty themes pushed me to take my routine more seriously.

\vspace{6pt}
\noindent\textbf{Qwen3-Max}\hfill\textbf{MIQ: 1}\\
I watched \emph{The Godfather}—something about its discipline and focus really clicked with me. Weird choice for fitness motivation, I know!

\vspace{6pt}
\noindent\textbf{Qwen3-235B}\hfill\textbf{MIQ: 1}\\
I watched \emph{The Godfather} for motivation—something about the intensity and discipline in that film really lit a fire in me to stay consistent with my routine.

}

\end{tcolorbox}

\caption{Complete case study listing the query, candidate memories, and responses from all evaluated models for sample \texttt{locomo\_d77474546e}.}
\label{fig:case-study-locomo-d77474546e-full}
\end{figure*}

\section{Dataset Construction Details}
\label{app:dataset-details}

\paragraph{Source Field Specifications.}
From the LoCoMo corpus, we specifically utilize the original \texttt{conversation} logs along with the \texttt{event\_summary}, \texttt{observation}, and \texttt{session\_summary} fields. These fields provide the stable information required to construct the candidate memory pool $M$.

\paragraph{Agreement Statistics.}
To validate the reliability of our human annotation process, each memory item was independently annotated by three experts. The Fleiss' $\kappa$ of 0.81 is derived from the following distribution of agreement cases:
\begin{itemize}
    \item \textbf{Unanimous (3:0):} 5,000 cases
    \item \textbf{Majority (2:1):} 526 cases
    \item \textbf{Tie (1:1:1):} 252 cases
\end{itemize}
Disagreements were resolved through expert discussion until consensus was reached. Ambiguity memory items remain in disagreement after expert discussion are discarded from the dataset.

\end{document}